\documentclass[manuscript,screen,nonacm]{acmart}

\AtBeginDocument{%
  }

\setcopyright{acmcopyright}
\copyrightyear{2024}
\acmYear{2024}
\acmDOI{XXXXXXX.XXXXXXX}

\acmPrice{15.00}
\acmISBN{978-1-4503-XXXX-X/18/06}

\usepackage{bm}
\usepackage{subcaption}

\usepackage{multirow}
\usepackage{mathtools,amssymb,euscript,yfonts,latexsym}
\usepackage[capitalise]{cleveref}
\usepackage{makecell}
\usepackage{soul}
\usepackage{dirtytalk}
\usepackage{amsmath}
\usepackage{threeparttablex}
\usepackage{color,colortbl}
\usepackage{textcomp}
\usepackage{cleveref}
\crefname{section}{§}{§§}

\usepackage{xparse}
\usepackage{nicematrix}
\usepackage{tikz}
\usetikzlibrary{calc}
\usepackage{lscape}
\usepackage{longtable}
\usepackage{lineno}

\definecolor{LightCyan}{rgb}{0.8,1,1}
\definecolor{Gray}{gray}{0.9}

\newcommand{\pb}[1]{\vspace{0.75ex}\noindent{\bf \em #1}\hspace*{.3em}}

\renewcommand{\paragraph}[1]{\vspace{1.25mm}\noindent\textbf{#1}}

\newcommand{\E}{\mathbb{E}}

\makeatletter
\usepackage{xspace}
\DeclareRobustCommand\onedot{\futurelet\@let@token\@onedot}

\ExplSyntaxOn
\NewExpandableDocumentCommand { \ValuePlusOne } { m } 
  { \int_eval:n { \int_use:c { c @ #1 } + 1 } }
\NewExpandableDocumentCommand { \Sec } { m } 
  { \fp_eval:n { secd ( #1 ) } }
\NewDocumentCommand { \Rot } { m }
  { 
    \hbox_to_wd:nn { 1 em }
      { 
        \hbox_overlap_right:n 
          { 
            \skip_horizontal:n { \fp_to_dim:n { 7 * cosd (\Angle) } } 
            \rotatebox{\Angle}{#1}
          } 
      } 
  }
\ExplSyntaxOff

\begin{document}
\title{Sora as a World Model? A Complete Survey on Text-to-Video Generation}

\author{Fachrina Dewi Puspitasari}
\affiliation{%
  \institution{University of Electronic Science and Technology of China}
  \country{China}
}
\email{puspitasari-dewi@outlook.com}

\author{Chaoning Zhang$\ast$} \thanks{\textbf{$\ast$Correspondence Author: Chaoning Zhang} (chaoningzhang1990@gmail.com)}
\affiliation{%
  \institution{University of Electronic Science and Technology of China}
  \country{China}
}
\email{chaoningzhang1990@gmail.com}

\author{Joseph Cho}
\affiliation{%
  \institution{Kyung Hee University}
  \country{South Korea}
}
\email{joyousaf@khu.ac.kr}

\author{Adnan Haider}
\affiliation{%
  \institution{University of Electronic Science and Technology of China}
  \country{China}
}
\email{adnanhaiderjaferi512@gmail.com}

\author{Noor Ul Eman}
\affiliation{%
  \institution{University of Electronic Science and Technology of China}
  \country{China}
}
\email{nooruleman3763@gmail.com}

\author{Omer Amin}
\affiliation{%
  \institution{University of Electronic Science and Technology of China}
  \country{China}
}
\email{oaminjr@gmail.com}

\author{Alexis Mankowski}
\affiliation{%
  \institution{University of Electronic Science and Technology of China}
  \country{China}
}
\email{alexis@mankowski.fr}

\author{Muhammad Umair}
\affiliation{%
  \institution{University of Electronic Science and Technology of China}
  \country{China}
}
\email{umair.analyzes@gmail.com}

\author{Jingyao Zheng}
\affiliation{%
  \institution{The Hong Kong Polytechnic University}
  \country{Hong Kong SAR}
}
\email{jingyao.zheng@connect.polyu.hk}

\author{Sheng Zheng}
\affiliation{%
  \institution{University of Electronic Science and Technology of China}
  \country{China}
}
\email{zszhx2021@gmail.com}

\author{Lik-Hang Lee}
\affiliation{%
  \institution{The Hong Kong Polytechnic University}
  \country{Hong Kong SAR}
}
\email{lik-hang.lee@polyu.edu.hk}

\author{Caiyan Qin}
\affiliation{%
  \institution{Harbin Institute of Technology Shenzhen}
  \country{China}
}
\email{qincaiyan@hit.edu.cn}

\author{Tae-Ho Kim}
\affiliation{%
  \institution{Nota Inc.}
  \country{South Korea}
}
\email{thkim@nota.ai}

\author{Choong Seon Hong}
\affiliation{%
  \institution{Kyung Hee University}
  \country{South Korea}
}
\email{cshong@khu.ac.kr}

\author{Yang Yang}
\affiliation{%
  \institution{University of Electronic Science and Technology of China}
  \country{China}
}
\email{yang.yang@uestc.edu.cn}

\author{Heng Tao Shen}
\affiliation{%
  \institution{Tongji University}
  \country{China}
}
\email{shenhengtao@tongji.edu.cn}

\renewcommand{\shortauthors}{Puspitasari et al.}


\begin{abstract}

The evolution of video generation from text, from animating MNIST to simulating the world with Sora, has progressed at a breakneck speed. Here, we systematically discuss how far text-to-video generation technology supports essential requirements in world modeling. 
We curate 250+ studies on text-based video synthesis and world modeling. We then observe that recent models increasingly support spatial, action, and strategic intelligences in world modeling through adherence to completeness, consistency, invention, as well as human interaction and control. 
We conclude that text-to-video generation is adept at world modeling, although homework in several aspects, such as the diversity-consistency trade-offs, remains to be addressed.

\end{abstract}





\maketitle

\section{Introduction}
On February 15\textsuperscript{th}, 2024, OpenAI introduced a new vision foundation model that can generate video from users' text prompts.
The model named Sora, which people call a video version of ChatGPT, has raised excitement from industries such as marketing~\cite{wallinger2024openais}, education~\cite{carballo2024openais}, and filmmaking~\cite{thomas2024sora} for promoting democratization and offering a shortcut to high-quality content creation.
OpenAI claimed that Sora, due to being trained on a large-scale dataset of text-video pairs, has impressive near-real-world generation capability.
This includes creating vivid characters, simulating smooth motion, depicting emotions, and provisioning detailed objects and backgrounds.

\pb{Survey Focus.}
Emerging from text-to-image generation models (denoted as \say{T2I}), text-to-video generation models (denoted as \say{T2V}) expand the technological features of the image counterpart to handle the temporal dimension in video.
Specifically, the generative model learns the approximation of the true input video data distribution with text as the condition.
On the other hand, world model defines a generative artificial intelligence (AI) model that understands general real-world mechanisms.
Naturally, for a T2V to be considered as world model, it requires skills beyond synthesizing video.
These world modeling skills includes, but not limited to, physics understanding, user visual comfort, and the logic of content composition.
Given the above assertions, we are interested in exploring \textit{how text-to-video generation models have come closer to being world models.}

\pb{Method.}
Our survey utilizes PRISMA~\cite{page2021prisma} framework.
We primarily collect resources from well-recognized publication venues, including, but not limited to, AAAI, ACL, CVPR, ECCV, ICCV, ICLR, IJCAI, NAACL, NeurIPS, ACM Multimedia, IEEE Transactions on Multimedia, and arXiv in November 2025. 
Given that text-to-video synthesis is a fast-growing study in the research community, we did not restrict the range of publication years.
Using the search keywords \say{text-to-video AND/OR (world model OR world simulator)}, we initially curated 400+ articles after briefly reviewing the fitness of the publication title.
We further devise several exclusion criteria, as follows:

\begin{itemize}
    \item articles that discuss video synthesizing not conditioned on text prompt,
    \item articles that discuss text-video relationship other than the generation task (e.g, retrieval, editing, captioning),
    \item articles that discuss world modeling unrelated to the generative vision task, and
    \item survey and review articles.
\end{itemize}

\noindent
and apply on the abstract and full article, we finally obtained a final list of 250+ papers.

\pb{Contribution.}
Following the recent surge in video generation, there have been a handful of surveys in T2V and/or world model.
Prior surveys 
discuss either T2V~\cite{waseem2025video, li2024survey, singh2023survey, xing2024survey, sun2024sora, liu2024sora} or world~\cite{ding2025understanding} model separately.
Another reviews both~\cite{zhu2024sora}, but centers the discussion on world modeling.
By contrast, we provide a comprehensive review of T2V technology from world modeling perspective.
Specifically, we first identify the critical components required for world modeling, then explore how the advancement in T2V has achieved them.
Our work also complements the existing survey in the generative AI models for text-to-text~\cite{zhang2023ChatGPT}, text-to-image~\cite{zhang2023text}, text-to-3D~\cite{li2023generative}, and text-to-speech~\cite{zhang2023audio}.
To summarise, our survey contributes the following to the research community:

\begin{itemize}
    \item through preliminary discussion on world modeling, we identify the substantial components required for a vision generative model to function as a world model,
    \item we provide technical discussion underpinning T2V from multiple aspects, including architecture, enabling technology, and other complementary elements that support its role as a world model.
\end{itemize}

\pb{Structure.}
We start by reviewing the foundations of world modeling within the perception-action system and how T2V contributes to it, from which we derive the essential aspects required by T2V to function as a world model (\cref{sec:foundation}).
We open the discussion on T2V by reviewing the historically important models as well as their core building blocks (\cref{sec:sota}).
With preliminary foundations on the world model and T2V, we then comprehensively discuss T2V from a world modeling perspective, which includes the discussion on technical enablers (\cref{sec:feature}), control and human-in-the-loop (\cref{sec:interaction}), dataset and evaluation (\cref{sec:datametric}), as well as applications (\cref{sec:application}).
Finally, based on these discussions, we summarize insights on how T2V can be improved in the future to become the true world model (\cref{sec:discussion}).

\section{Preliminary on World Models}\label{sec:foundation}
\subsection{Definitions of World Model}

\noindent
\begin{minipage}{0.45\linewidth}
Imagine a waiter pouring a glass of orange juice: through experience, he intuitively determines the precise level to stop, preventing the liquid from overflowing.
This judgment arises from his understanding of factors such as the jug’s tilt angle, the juice level in the glass, and the experience accumulated through repeated practice.
The waiter's intuitive sense demonstrates an internal understanding of how actions lead to outcomes.
Such predictive reasoning illustrates the principle of a world model.
\end{minipage}
\hfill
\begin{minipage}{0.5\linewidth}
\includegraphics[width=\linewidth]{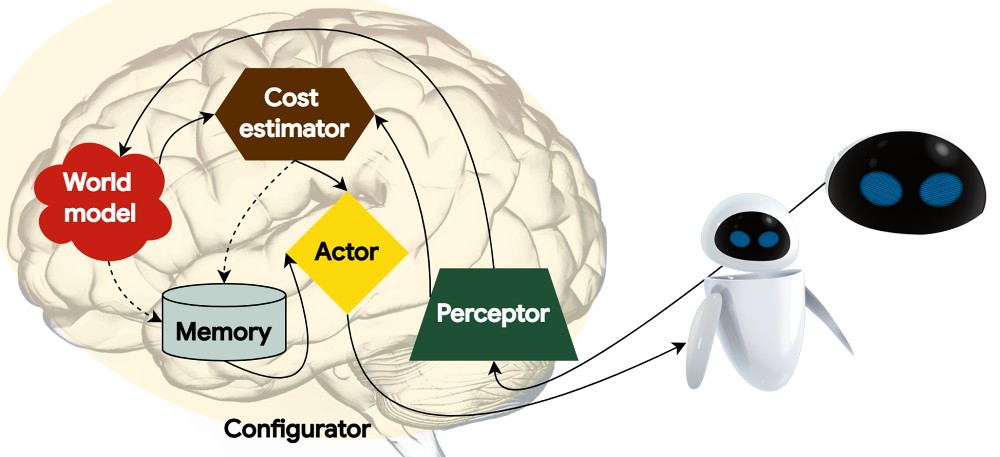}
\captionof{figure}{World model inside the perception-action system. The world model is part of the system brain that connects the internal representation to the external world (adapted from~\cite{lecun2022path}).}
\label{fig:perceptact}
\end{minipage}

\pb{Mathematical Definition.}
\emph{World model} conceptualizes the interaction between an agent and its environment~\cite{damm2011does}.
\say{World} refers to the environment where the agent resides and \say{model} refers to the vision of the world carried by the agent~\cite{ha2018world}.
The agent constructs a world model with only a subset of world elements that are relevant to the task at hand~\cite{lecun2022path}.
The idea of world model originates from the concept of mental model in psychology~\cite{fox2022human}, in which internal and external systems interact with each other.
Mathematically, world model $M$ can be defined as follows~\cite{damm2011does}:

\begin{equation}
M = \left(V, S, s_0, A, F_A, F_S\right)
\end{equation}

\noindent
where $V$, $S$, $s_0$, $A$, $F_A$, and $F_S$ refer to the environment variable, all possible outcome states, initial state, the action taken, the valuation function of responses variable, and the valuation function of control variables, respectively.

\pb{World Model In Intelligent Systems.}
World model is a substantial component of the agents' intelligence system, which is optimized to improve the decision-making skills~\cite{suomala2022human} through interaction with the environment~\cite{sutton1998reinforcement}.
The system, referred to as \underline{perception-action system} (Fig.~\ref{fig:perceptact}), consists of six interconnected modules (i.e., perceptor, world model, cost estimator, actor, short-term memory, and configurator)~\cite{lecun2022path}.
Perceptor estimates the current state of the world based on input, and world model thinks about the possible future state.
The discrepancies between these two modules are examined by cost estimator, which then becomes an input for the actor to find an optimal action that minimizes this cost in the future sequence.
Information about the world state, predicted state, and their cost is stored in the short-term memory that obeys temporal logic~\cite{allen1983planning}.
Finally, configurator regulates the information flow in the whole system.

\subsection{Roles of World Model in Perception-Action System}

\begin{figure}[!htb]
     \centering
     \includegraphics[width=\linewidth]{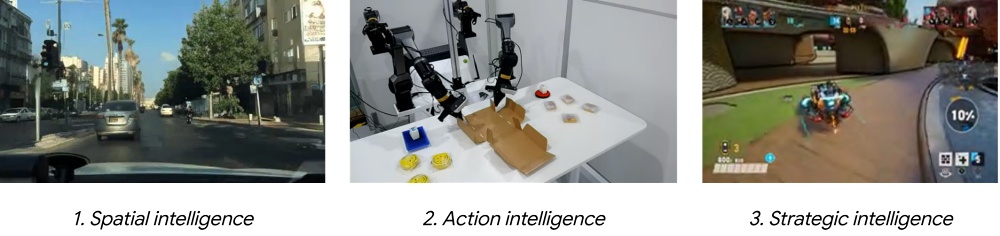}
     \caption{Cases of perception-action system where the world model performs spatial~\cite{mousakhan2025orbis}, action~\cite{liao2025genie}, and strategic~\cite{kanervisto2025world} intelligences.}
     \label{fig:intelligence}
\end{figure}

Perception-action system serves as the physical actualization of AGI.
The world model is a part of AGI, acting as the prefrontal cortex and hipopocampus which executes two core functions: prediction and planning, and memory storage.
In this section, we discuss three core roles of world model in the perception-action system: spatial intelligence, action intelligence, and strategic intelligence (Fig.~\ref{fig:intelligence}).

\pb{Spatial Intelligence.}
Spatial intelligence refers to the agent's ability to see, process, and navigate through various spatial layouts.
For instance, a home robot adapting to a customer's house layout, or a city ego vehicle that adapts to rural roads.
Given the abundance of spatial layout combinations, it is impossible to pre-define everything during training.
Thus, the role of world modeling is needed to circumvent this combinatorial knowledge problem.
Spatial intelligence is commonly performed as a scene prediction task, where given a single view of the spatial layout, the world model can generate its prediction on the 360$^\circ$ view~\cite{koh2021pathdreamer}.
Spatial intelligence entails four substantial aspects that must be adhered to by the world model:
\begin{itemize}
    \item \textit{Visual fidelity}: ensuring high-quality visual construction~\cite{alonso2024diffusion}.
    \item \textit{Temporal consistency}: acceleration~\cite{zhao2025drivedreamer4d} of all moving objects that determine the occlusion hierarchy~\cite{yang2025instadrive} and visual condition~\cite{lu2024wovogen}.
    \item \textit{Structural stability}: handling of corner cases and viewpoint responses to unrealistic agent behavior~\cite{mousakhan2025orbis, po2025long}.
    \item \textit{Commonsense plausibility}: adherence to world law (e.g., traffic signals)~\cite{chang2025seeing}.
\end{itemize}
The present-day world model in the perception-action system has considered these crucial aspects to ensure seamless world modeling.
For example, this is achieved by injecting spatial priors (e.g., point cloud and depth map, perspective map)~\cite{yan2025streetcrafter, mei2024dreamforge}. 
Some world models even generate spatial priors (e.g., 3D occupancy map, LiDAR map)~\cite {wang2024occsora, zyrianov2025lidardm} to be utilized as the intermediate stage for the scene generation~\cite{lu2024wovogen, lu2025infinicube}.

\pb{Action Intelligence.}
Action intelligence specifies how agent behaves in task execution.
A naive example is an assistant robot that can mimic human activity, performing behavioral cloning.
Action intelligence undergoes an open-ended objectives problem that may consist of not only \textit{what}, but also \textit{how} an activity is performed.
Besides, it also suffers from the scarcity (even more scarce than spatial intelligence) of data for training a generalist agent.
World models, with their strong generative capabilities, offer a promising direction for addressing these challenges~\cite{liao2025genie}.
To perform action intelligence, the world model needs to comply with two fundamental requirements:
\begin{itemize}
    \item \textit{Embodiment}: reducing the perception–action gap between visual examples and real-world execution~\cite{li2025multimodal, liang2024dreamitate}.
    \item \textit{Physical plausibility}: obeying physics law in real world~\cite{yang2025physworld, soni2025videoagent}.
\end{itemize}
Some models depend on high-fidelity generated vision and state transition to learn these key elements~\cite{ren2025videoworld}.
Some others rely on external data as generation conditions, such as sensorimotor signals (e.g., haptic forces, EMG, body pose, and gaze tracking)~\cite{li2025multimodal, bjorck2025gr00t}, and robot trajectories~\cite{wu2024ivideogpt}.
Other models decouple action and vision modals~\cite{zhu2025unified}, rely on a digital twin to transfer the action knowledge into new environments~\cite{fang2025rebot}, or attach an action adapter to a strong world model~\cite{rigter2024avid} to get the benefit from large-scale action-free data and exclusively curated behavior data.
World model as action intelligence is also utilized as policy in reinforcement learning~\cite{tseng2025scalable, liang2025video}.
It is also further enhanced by an interim reflection mechanism that validates the generated content before it is translated into robot action~\cite{chi2025empowering}.

\pb{Strategic Intelligence.}
Strategic intelligence refers to agents' ability to perform long-horizon reasoning, opponent modeling, and rule understanding.
Strategic intelligence encompasses all elements in the world that the agent lives in. 
This includes the environment and all the actors, including non-player characters.
Game is an example of where strategic intelligence is needed.
A pleasurable gaming experience needs the integration of four key aspects: task leveling, game mechanics, players, and randomness~\cite{kanervisto2025world}.
Deriving from this, there are two key requirements for a world model to perform strategic intelligence:
\begin{itemize}
    \item \textit{Diversity}: generating diverse scenarios that fulfill the users' divergent thinking~\cite{bruce2024genie}.
    \item \textit{Interaction mechanics}: ensuring and sustaining~\cite{he2025pre} coherence between users' inputs and the generated content~\cite{yang2024playable}, as well as demonstrating causality from the interaction~\cite{kim2020learning}.
\end{itemize}
Interactive behaviors can be integrated into generated content by learning latent action representations from gameplay data~\cite{bruce2024genie} or through imagination-based training~\cite{hafner2025training}.
Decoupling domain modeling from action learning further enables the synthesis of novel game scenarios while maintaining realistic action control~\cite{yu2025gamefactory, ye2025yan}.
User interaction is typically expressed via general atomic control signals, such as keyboard bindings in game-generation settings~\cite{che2024gamegen, ye2025yan}.
Diversity can be introduced through reinforcement learning frameworks that cultivate agents with diverse skill sets, whose action trajectories are subsequently used to train generative models~\cite{valevski2024diffusion}.
Moreover, several approaches employ learned dynamics engines to model state transitions explicitly, allowing the system to capture interaction causality~\cite{kim2020learning, hafner2025training, bruce2024genie}.
Finally, memory mechanisms are often incorporated to support long-horizon reasoning and planning~\cite{wu2025video, kim2020learning, yang2024playable}.


\subsection{Why Text-to-Video Generation Matters for World Modeling}

In perception–action systems, visualization is a core mechanism for interacting with the environment. 
Vision, in particular, serves as an efficient medium for information transmission~\cite{kepes1995language}.
Given that effective simulation requires both spatial and temporal understanding~\cite{oliveri2009spatial, rohenkohl2014combining}, most existing systems rely on video for learning, sourced from real-world recordings~\cite{finn2017deep} or video games~\cite{ha2018recurrent}. 
However, collecting video data that sufficiently covers the breadth of real-world scenarios remains impractical.
While video games can simulate diverse environments, they often rely on controlled scenarios that hinder real-world generalization.
Although internet videos are abundant, they are typically noisy and lack critical properties for effective learning, such as skill development and realistic challenges~\cite{narayanasamy2005distinguishing}.
T2Vs overcome these limitations by generating diverse, task-specific environments that serve as agents' training playground~\cite{he2025pre}.

\subsection{Critical Aspects of Text-to-Video for World Modeling}
%


Generated video content from T2V serves as an imagined world, similar to the environments constructed in fiction and video games.
Such worlds are characterized by three core properties: invention, completeness, and consistency~\cite{wolf2014building}.
\textbf{Completeness} consists of comprehensiveness and cohesion~\cite{rudy2014introduction} dimensions that correspond to content alignment and richness, and scene compatibility, respectively.
In perception-action systems, completeness is manifested as both \textit{visual fidelity} and \textit{interaction mechanics}.
\textbf{Consistency} measures whether generated content is plausible, non-contradictory, and adheres to the rules of the world~\cite{lessa2017world}.
It includes entity, composition, and world consistency.
Entity consistency concerns the stable appearance of objects across space and time, composition consistency captures coherent spatial relationships and interactions, and world consistency enforces physical and semantic plausibility.
In the perception-action system, consistency is translatable as \textit{temporal consistency}, \textit{structural stability}, and \textit{semantic and physical plausibility}.
\textbf{Invention} is creativity constrained by internal consistency~\cite{wolf2014building}. 
At the same time, invention enables extensibility, supporting expansion and allowing new scenes, actions, and viewpoints to emerge without contradiction. 
In a perception–action system, this extensible yet rule-consistent structure manifests as \textit{diversity} in generation.

\section{Text-to-Video as World Modeling}\label{sec:sota}

Since the emergence of Sora, T2Vs have increasingly been positioned as candidates for visual world modeling. 
Both open-source and proprietary models are now competing to produce highly realistic videos while offering flexible and fine-grained user control. 
In this section, we identify the current top-performing T2V models based on the public video generation leaderboards and review their key design choices and enabling technologies. 
For closed-source models with limited technical disclosure, we infer their underlying mechanisms from related research outputs and technical reports.

\begin{figure}[!htb]
     \centering
     \includegraphics[width=.9\linewidth]{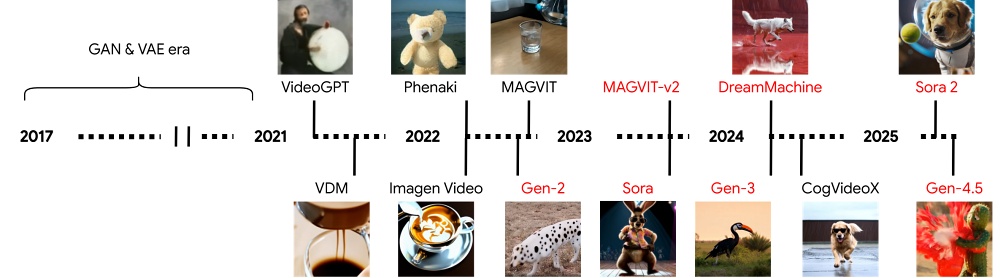}
     \caption{Milestone of several historically substantial text-to-video generation models. Models in \textcolor{red}{red} are proprietary.}
     \label{fig:history}
\end{figure}

\subsection{Historical Development of Text-to-Video Generation Models}

Much of the technology utilized in the current top-performing T2Vs originates from early T2Vs.
Here, we discuss several historically important models (Fig.~\ref{fig:history}) that have become the inspiration for present-day world-model-like T2Vs.

\pb{Foundational Text-to-Video Generation Models.}
\underline{Video Diffusion Model} (VDM)~\cite{ho2022video} is the first extension of T2I to T2V.
The model inherits the 3D U-Net architecture from T2I to generate videos with a fixed number of frames.
The key innovation of VDM is \textbf{factorized space-time attention} on 1D+2D latents over pseudo-3D convolution that enables seamless image-video joint training.
\underline{Stable Video Diffusion}~\cite{blattmann2023stable} (SVD) then improves the training paradigm into three stages: T2I pretraining, video pretraining, and
high-quality video finetuning.
Further, to enable denser frame generation during inference, VDM proposes reconstruction-guided sampling, where the intermediate output is checked first to see whether it satisfies the input condition. 
Meanwhile, \underline{Imagen Video}~\cite{ho2022imagen} presents techniques for high-fidelity video generation over a fixed number of frames.
The essential innovation of Imagen Video is a \textbf{cascaded diffusion framework}, where the base T2V model is empowered by interleaved spatial and temporal video super-resolution.
This model also implements factorized space-time attention as well as progressive distillation for fast training.
Finally, \underline{Phenaki}~\cite{villegas2022phenaki} pioneers arbitrary-length video generation.
The key idea proposed by Phenaki is C-ViViT, which employs \textbf{causal attention for autoregressive-like generation}.
Further, Phenaki utilizes masked bidirectional transformer to enable efficient sampling for long video generation.

\pb{Foundational Transformer-based Models.}
\underline{VideoGPT}~\cite{yan2021videogpt} is the first T2V that explores a text-generation model instead of a diffusion model for video generation.
The key finding of VideoGPT is that the potential of \textbf{transformer-based architecture that enables world modeling}.
Meanwhile, \underline{MAGVIT}~\cite{yu2023magvit} proposes 3D VQ-VAE to quantize video into discrete tokens suitable for text-generation models.
The essential innovation of MAGVIT is enabling \textbf{multi-task generation} with masked token modeling that utilize interior condition as a learnable task router.
Further, \underline{MAGVIT-v2}~\cite{yu2023language} constructs a more powerful video tokenizer with a causal 3D CNN to support the implementation of a text-generation model for generating video.
The study shows that this technical improvement enables the \textbf{transformer-based language model to perform better than the U-Net-based diffusion model} for video generation.

\pb{Early Proprietary Models.}
\underline{Runway Gen-2}~\cite{runway2023gen} pioneers the widely accessible T2V as a \textbf{video generation tool for the public}.
Gen-2 also provides a wide range of controls and input modalities for users to perform various video-related tasks.
Impressively, despite being an early proprietary model, Gen-2 is capable of generating realistic video.
\underline{Runway Gen-3}~\cite{runway2024gen} further improves Gen-2 performance in terms of \textbf{visual fidelity and consistency}.
Equipped with the more advanced controls, Gen-3 generates more highly realistic subjects compared to Gen-2.
Moreover, \underline{Dream Machine}~\cite{luma2024dream} establishes a benchmark for proprietary models in terms of \textbf{physical realism} and motion.

\pb{Top-Performing World-Modeling-Aware Models.}
\underline{Runway Gen-4.5}~\cite{runway2025introducing} is currently among the highest-ranked T2Vs and has been adopted by numerous industry leaders. 
Beginning with Gen-1 in 2023, which was built upon a latent diffusion backbone, Runway has evolved into its latest version, offering richer control mechanisms and enhanced in-context generation capabilities. 
\underline{Veo 3}~\cite{google2025introducing} is among the top-performing T2Vs that support native audio generation alongside video synthesis. 
Similar to Runway, Veo 3 emphasizes 
audio-conditioned generation, long video extension, and improved character consistency.
\underline{Sora 2}~\cite{openai2025sora} is introduced by OpenAI one year after the release of Sora.
The model is believed to retain Sora’s Diffusion Transformer (DiT) backbone while strengthening physical plausibility. 
\underline{Minimax Hailuo}~\cite{minimax2025minimax} offers user video generation with complicated physical simulation due to its noise-aware compute redistribution that allows for highly efficient computation.
\underline{Seedance}~\cite{gao2025seedance} optimizes text and vision representation individually through separated branches to improve multimodal alignment.
\underline{Lumiere}~\cite{bar2024lumiere} introduce \textbf{space-time U-Net where the temporal dimension is generated in a single-stream} generation, followed by spatial super-resolution to prioritize temporal continuity.
The hype of T2V also promotes the deployment of other proprietary models, such as PixVerse~\cite{pixverse2025pixverse}, Pika 1.5~\cite{pika2025pika}, Dreamina~\cite{dreamina2025dreamina}, Hailuo~\cite{hailuo2025hailuo}, and  Kling~\cite{kling2025kling}. 
On the other hand, open-source models also increasingly improve.
\underline{HunyuanVideo}~\cite{kong2024hunyuanvideo} released by Tencent in 2025 demonstrates competitive performance with closed-source counterparts in terms of temporal consistency, dynamics, and physical plausibility. 
\underline{CogVideoX}~\cite{yang2024cogvideox}, developed by Tsinghua University, is an open-source T2V that achieves performance comparable to leading closed-source systems. 
\underline{Mochi}~\cite{genmo2024mochi} proposes \textbf{asymmetric DiT that performs text and vision processing in different streams} and joins them with multimodal self-attention, optimizing each modality and expediting inference speed.
\underline{Vchitect}~\cite{fan2025vchitect} also implement conceptually similar architecture to Mochi.
\underline{Wan}~\cite{wan2025wan} offers fast computation and maintains temporal continuity due to its \textbf{feature caching of preceding frames}.
We present the performance of these models across all evaluation dimensions in Table~\ref{tab:score}.

\subsection{Core Building Blocks}
\pb{Text Encoders.}
A text encoder encodes the raw text prompt into embedding representations that become input for the vision model as a condition.
Since its introduction in 2017, text encoders utilized by T2V models have followed the evolution of text modeling.
For instance, recurrent-based text models (e.g., RNN, LSTM, and GRU) are more common in early-era T2V models, while transformer-based models (e.g. T5~\cite{raffel2020exploring} and BERT~\cite{devlin2019bert}) are more prevalent in modern T2V models.
Recent T2V models utilize CLIP~\cite{radford2021learning} text encoder for its remarkable global contextual understanding.

\begin{figure}[!htb]
     \centering
     \includegraphics[width=.8\linewidth]{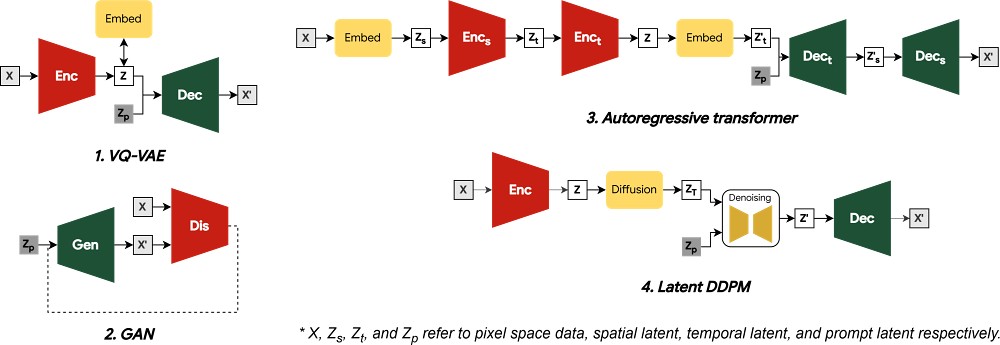}
     \caption{Generative vision backbone in text-to-video generation models.}
     \label{fig:vision}
\end{figure}

\pb{Vision Model.}
A vision model converts input (e.g., image or noise) and condition (e.g., text embedding) into an output video.
T2V models mainly inherit the main architecture from T2I models.
These vision backbones are generally grouped in four architectures: VAE, GAN, autoregressive transformer, and latent DDPM (Fig.~\ref{fig:vision}).

\begin{figure}[!htb]
     \centering
     \includegraphics[width=.9\linewidth]{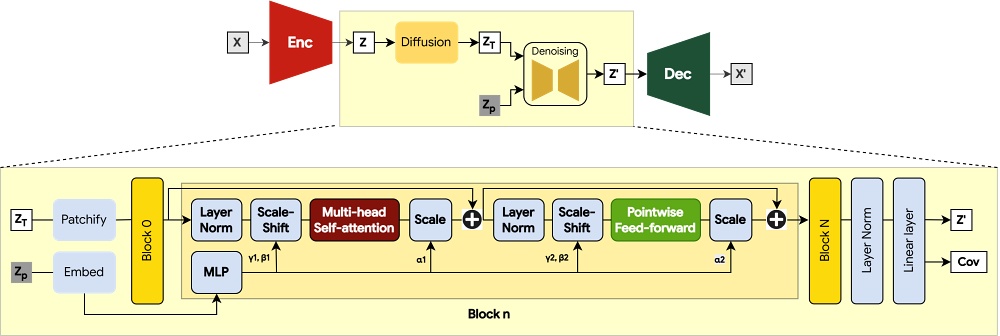}
     \caption{The architecture of Diffusion Transformer, which combines a transformer with latent DDPM (adapted from~\cite{peebles2023scalable}).}
     \label{fig:dit}
\end{figure}

\begin{itemize}
    \item \underline{VAE and GAN.}
    \textit{VAE}~\cite{kingma2013auto} is a probabilistic modeling based on variational inference of $q(z \mid x)$ that approximates the posterior $p(x \mid z)$ by maximizing the Evidence Lower Bound loss. 
    \begin{equation}
    \mathcal{L}_{\text{VAE}}
    = 
    - \mathbb{E}_{q_{\phi}(z \mid x)} \left[ \log p_{\theta}(x \mid z) \right]
    + \mathrm{KL}\!\left( q_{\phi}(z \mid x) \,\|\, p(z) \right).
    \end{equation}
    Meanwhile, GAN~\cite{goodfellow2014generative} works according to the Minmax game principle between generator $G$ and discriminator $D$, where the former generates samples, and the latter tries to distinguish between real and generated samples.
    \begin{equation}
    \min_{\theta_g} \max_{\theta_d} 
    \; \mathbb{E}_{\mathbf{x} \sim p_{\text{data}}} \!\left[ \log D_{\theta_d}(\mathbf{x}) \right]
    + 
    \mathbb{E}_{\mathbf{z} \sim p(\mathbf{z})} \!\left[ \log \left( 1 - D_{\theta_d}(G_{\theta_g}(\mathbf{z})) \right) \right].
    \end{equation}
    This training paradigm encourages GAN to produce diverse high-quality samples, making GAN-based T2V models often utilized for story visualization (e.g., StoryGAN~\cite{li2019storygan}).
    Further, VQ-VAE and VQ-GAN allow T2V models (e.g., GODIVA~\cite{wu2021godiva}) to perform generation process in the discrete latent space to solve the prohibitive computational cost from pixel-space generation. 

    \item \underline{Autoregressive Transformer.}
    Learning from how a discrete latent space has effectively facilitated text-vision joint learning, some modern T2V models (e.g., VideoPoet~\cite{kondratyuk2023videopoet}) synthesize video using a transformer.
    This idea is pioneered by Phenaki through its C-ViViT architecture.
    MAGVIT-v2 later shows that this representation discretization offers several benefits compared to the diffusion model, such as supporting multiple modes of communication, speeding up (de)compression, and contextual understanding.

    \item \underline{Latent DDPM.}
    Diffusion model, specifically Stable Diffusion~\cite{rombach2022high}, is the most popular vision model utilized by contemporary T2Vs.
    Stable diffusion performs generation from input data $\textbf{x}$ in continuous latent space encoded by VAE.
    Diffusion model or DDPM~\cite{ho2020denoising} offers a better balance between diversity and fidelity compared to GAN~\cite{dhariwal2021diffusion}.
    DDPM working mechanism entails two major processes, forward and backward.
    \textit{Forward} process converts input data $\textbf{z}_{0}$ into noise \( \textbf{z}_{T} \sim \mathcal{N} \)(\textbf{0}, \textit{\textbf{I}}) by gradually adding an infinitesimal amount of noise for a $T$ timesteps.
    Meanwhile, the \textit{backward} process performs generation by iteratively denoising data \( \textbf{z}_{t-1} \) from \( \textbf{z}_{t} \) using a neural network \( \boldsymbol{\epsilon}_{\theta}(\textbf{z}_{t}, t) \) which is optimized through the following objective function:
    
    \begin{displaymath}
    L^{simple}_{t}=\E_{t\sim[1,T],\textbf{x}_{0},\boldsymbol{\epsilon}_{t}}\left[\parallel\boldsymbol{\epsilon}_{t}-\boldsymbol{\epsilon}_{\theta}(\sqrt{\overline{\alpha}_{t}}\textbf{x}_{0} + \sqrt{1-\overline{\alpha}_{t}}\boldsymbol{\epsilon}_{t}, t) \parallel^2\right]
    \end{displaymath}
    
    \noindent
    Modern DDPM allows users to control the generation process by inserting prompts such as text, audio, and localization marks.
    The classifier-free diffusion guidance handles all of these extra inputs. 
    It is trained along with the DDPM neural network during the reverse denoising process~\cite{ho2022classifier}.

    \item \underline{Diffusion Transformer.}
    Diffusion Transformers (DiT)~\cite{peebles2023scalable} serve as the dominant backbone in contemporary T2Vs. 
    By combining the generative strengths of latent DDPM with the generalizability of transformers, DiT provides a highly scalable architecture for video generation. 
    This scalability enables the synthesis of longer, higher-quality videos with improved alignment to user queries.
    The core architectural innovation of the DiT lies in replacing the convolutional U-Net in latent DDPMs with a transformer-based backbone (Fig.~\ref{fig:dit}).
    DiT operates on a sequence of $N$ patch tokens obtained by linearly projecting the VAE-encoded latent representation $z_0$. 
    In addition to visual tokens, the transformer incorporates conditioning signals such as diffusion timesteps, class labels, or text embeddings. 
    These conditioning inputs modulate the network through adaptive layer normalization, where scale-shift ($\gamma$ and $\beta$) parameters are learned directly from the conditioning embeddings. 
    DiT further introduces residual scaling to stabilize training, with scaling parameters ($\alpha$) initialized to zero. 
    After a stack of transformer blocks and a final normalization layer, a linear projection produces the predicted noise (and optionally covariance), which is used in the reverse denoising process to generate the output image.

\end{itemize}

\begin{figure}[!htb]
     \centering
     \includegraphics[width=.8\linewidth]{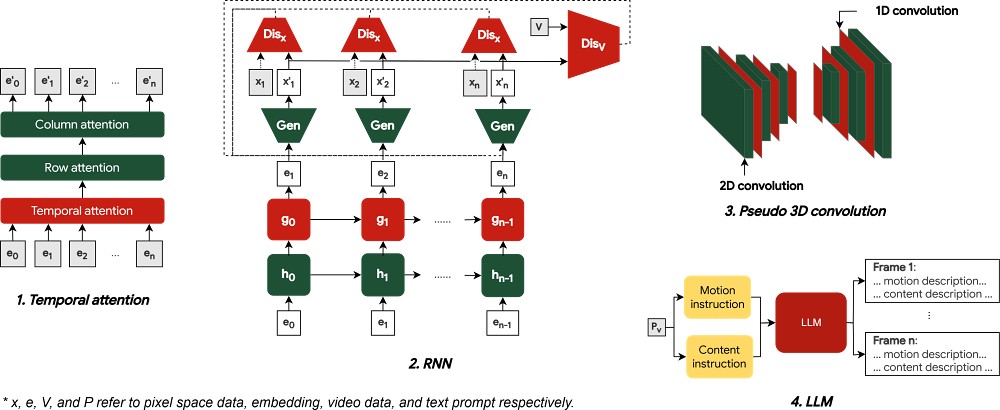}
     \caption{Temporal handler used to align frames in text-to-video generation models.}
     \label{fig:temporal}
\end{figure}

\pb{Temporal Handler.}
Temporal handler complements the vision model by learning the inter-frame visual dynamics.
This function exists in a range of mechanisms: temporal attention, RNN, pseudo-3D convolution and attention, and LLM (Fig.~\ref{fig:temporal}).

\begin{itemize}
    \item \underline{RNN.}
    Recurrent neural networks, such as 
    LSTM~\cite{hochreiter1997long} and GRU~\cite{cho2014learning} are commonly integrated into GAN-based vision models (e.g., StoryGAN).
    RNN takes text representation, noise, and its previous hidden state to produce a new hidden state, which becomes the input to the generator.
    To ensure both temporal consistency and content diversity, auxiliary modules like a hint generator 
    connects inter-frame output~\cite{kim2020tivgan,li2019storygan,lei2020mart}.

    \item \underline{Temporal Attention.}
    Adding a temporal attention that operates on time dimension is the most straightforward way of incorporating temporality into T2I models.
    Temporal layers are commonly found in autoregressive and VQ-VAE vision models.
    Several implementations of temporal layers are found in Phenaki with causal attention, CogVideo~\cite{hong2022cogvideo} with temporal attention, GODIVA~\cite{wu2021godiva} with 3D-sparse attention, and VideoPoet~\cite{kondratyuk2023videopoet} that utilizes causal attention on regressive language model.

    \item \underline{Pseudo-3D Convolutions.}
    Pseudo network inflation is a familiar technique for incorporating temporal dimensionality into the diffusion model, such as Make-A-Video~\cite{singer2022make} and the majority of latent DDPM-based T2Vs.
    Here, a 1D convolution (temporal) layer is attached after every 2D convolution (spatial) layer.
    Through the pseudo-3D convolution layer, the input video of shape $batch \times frames \times channels \times height \times width$ is processed by the spatial layer as $(b \times f) \times c \times h \times w$ and the temporal layer as $b \times c \times f \times h \times w$.
    This separable convolution~\cite{chollet2017xception} technique offers multiple benefits: circumventing expensive computation, 
    preserving the 2D backbone knowledge while updating the temporal parameter, and sharing knowledge with all of the input elements.
    In addition to pseudo-3D convolution, inflating T2I to T2V can also be enriched with noise prior correlation~\cite{ge2023preserve}.

    \item \underline{Pseudo-3D Attention in DiT.}
    Conceptually similar to pseudo-3D convolutions, pseudo-3D attention layer also offers cheap temporal integration.
    Pioneered by VDM, this layer
    leverages sinusoidal positional embedding to supply frame indices information.
    Some examples of practical implementations include Gentron~\cite{chen2024gentron} and Open-sora~\cite{zheng2024open} that utilize lightweight temporal self-attention, Latte~\cite{ma2024latte} and MotionZero~\cite{su2023motionzero}, which interleave spatial and temporal attention, and Runway that attaches a temporal extension module in a residual connection and attention block.
    To ensure that the latent representations preserve the temporal continuity, CogVideoX leverages 3D-VAE~\cite{zheng2024open}, while another improves it with leading padding to encode temporal causality~\cite{yu2023language}.
    Further, to handle temporal inconsistency in fast motion, CogVideoX utilizes 3D full attention~\cite{yang2024cogvideox}, while another supplies motion conditioning~\cite{chen2024gentron}.
    Finally, it is worth noting that these models achieve proper temporal modeling by treating T2V as the generation of \say{image as $T$ frames video} instead of \say{video as a sequence of $T$ images}~\cite{menapace2024snap}.
\end{itemize}

\subsection{Towards World Modeling Backbone}
\pb{Scalable and Generalizable Learning.}
Training on an immense number of high-resolution images and videos, while improving generalizability, is highly prohibitive.
To circumvent this issue, MDTv2~\cite{gao2023masked} and DiffPhy~\cite{zheng2023fast} subsequent works propose masked training, where it learns to predict the score of unmasked patches and reconstruct the masked patches simultaneously.
PIXART-$\alpha$~\cite{chen2023pixart} decouples pixel learning and text-image learning from the standard unified training, making intra- and inter-modal learning less complex and thus more efficient.
AnimateLCM~\cite{wang2024animatelcm} decouples image prior and motion prior to expedite learning.
Runway decomposes video generation into content (e.g, appearance) and structure (e.g., geometry and dynamics)~\cite{esser2023structure}.
Structural knowledge can be derived using structural encoders, while content is derived from CLIP image embedding that is sensitive to semantic properties~\cite{li2025bindweave}.
Finally, pioneered by VDM, nearly all current T2Vs (e.g., Gentron~\cite{chen2024gentron} and Latte~\cite{ma2024latte}) employ joint image-video training to get the advantage of high visual quality and high temporal consistency.

\pb{Efficient Sampling.}
The iterative sampling process during the denoising step is one of the bottlenecks of DiT.
Consistency model~\cite{song2023consistency, song2023improved} offers a remedy to this limitation.
Specifically, consistency models enable sampling high-quality data in a single step or \textit{n}-steps where \textit{n} is much smaller than the original iteration steps.
To date, the effort to compress the sampling step has developed into various approaches.
Some approaches perform modification by replacing denoising with rectified flow that directly learns the velocity field to transport noise to data~\cite{esser2024scaling}, replacing DiT with an interpolant transformer that takes advantages from denoising, rectified flow, and others~\cite{ma2024sit}.
Some of which are utilizing the probability density function of the signal-to-noise ratio to schedule sampling~\cite{yao2024fasterdit} and weighted caching of representation at each step with error optimization~\cite{qiu2025accelerating}.
Other approaches explore the asymmetric sampling, such as sampling prioritizing areas that have not yet converged~\cite{wang2025closer}, adjusting the model shape at each timestep~\cite{zhao2025dynamic}, and treating the sampling step as a mixture-of-depth~\cite{shen2024mddit}.

\section{World-Modeling-Enablers in Text-to-Video Generation}\label{sec:feature}

\begin{figure}[!htb]
     \centering
     \includegraphics[width=.9\linewidth]{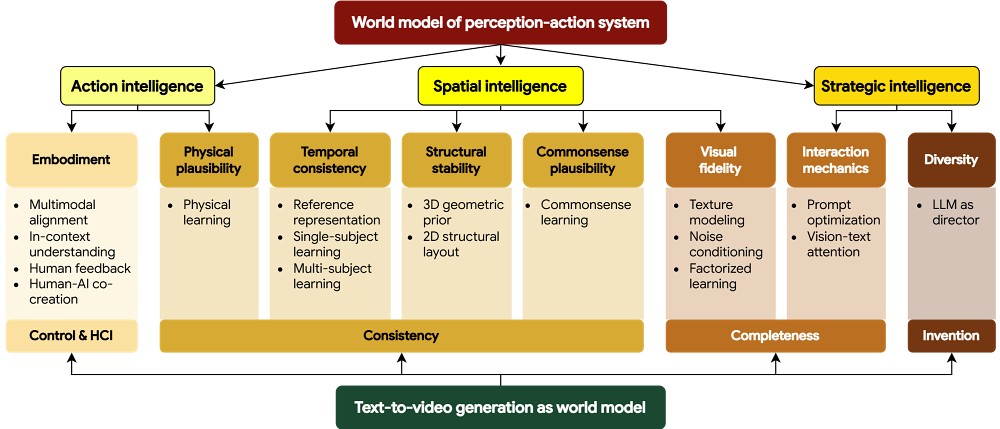}
     \caption{The relationship between essential components for world model of the perception-action system and substantial components of text-to-video generation as world model.}
     \label{fig:t2v}
\end{figure}

T2V as a world model needs to adhere to the specification on world model for perception-action system (Fig.~\ref{fig:t2v}).
Accordingly, this section reviews the key techniques adopted in recent T2V models to address these additional requirements.

\subsection{Visual Fidelity}
Visual fidelity is a foundational requirement for any vision generation task.
As T2V generation builds upon T2I models, it naturally inherits many image-generation challenges, such as preserving fine-grained textures and handling non-zero signal-to-noise ratios.

\pb{Texture Modeling.}
Texture plays a crucial role in perceived image quality, as it conveys fine-grained visual details. 
However, texture information is often suppressed when images are compressed into latent representations. 
Cascaded generation, as pioneered by Imagen Video, offers an effective solution to this limitation.
Specifically, models like VideoLDM~\cite{blattmann2023align}, Waver~\cite{zhang2025waver}, and Show-1~\cite{zhang2025show} first generate low-resolution frames in pixel space to preserve fine details, and subsequently enhance them via latent-space super-resolution. 
In addition, texture information can be explicitly introduced as a conditioning signal during generation through semantic-texture consistency training~\cite{yu5137309optimizing}.
To alleviate the difficulty of reconstructing good representations, regularization based on distillation of external high-quality representations can guide model learning~\cite{yu2024representation}.

\pb{Noise Conditioning.}
Diffusion models often suffer from reduced visual fidelity due to a mismatch between training and inference noise schedules.
Residual signals at the terminal training timestep hinder generalization to pure Gaussian noise during sampling.
This issue can be mitigated by aligning the two processes, for example, Emu Video~\cite{girdhar2023emu} scales the noise schedule to zero at the final timestep.
On the other hand, to alleviate train-test discrepancies in noise scheduling, ART$\cdot$V~\cite{weng2024art} performs noise augmentation.


\pb{Factorized Learning.}
Jointly optimizing spatial and temporal dimensions remains a core challenge in T2V generation.
Disentanglement provides an effective remedy by simplifying the learning problem.
For example, it can be realized through spatiotemporal decomposition in prompt level~\cite{wang2025identity}, latent space level~\cite{jiang2023text2performer}, and architectural level~\cite{liu2025separate}.
HiGen~\cite{qing2024hierarchical} implements hierarchical spatiotemporal learning.
Similarly, MotionBooth~\cite{wu2024motionbooth} additionally incorporates subject region loss and video preservation loss.
In hierarchical learning approaches, models are typically trained in a sequential order, such as T2I $\rightarrow$ T2V or T2V $\rightarrow$ T2I.
However, differences in noise sampling mechanisms between image and video diffusion often introduce visual inconsistencies.
To address this issue, recent works pursue \emph{pseudo-simultaneous} learning of spatial and temporal representations.
To reduce flickers and generate high visual quality, DSDN~\cite{liu2023dual} employs dual-stream diffusion and EVS~\cite{su2025encapsulated} encapsulates T2I with T2V in a T2I$\xrightarrow{}$T2V$\xrightarrow{}$T2I manner.
On the other hand, DCM~\cite{lv2025dual} performs dual-expert consistency modeling and cross-dimensional losses, and MotionShot~\cite{liu2025motionshot} learns semantic and morphological motion hierarchically.

\subsection{Diversity}
A video must exhibit not only temporal consistency but also sufficient visual diversity and progression to convey a story.
However, generating diverse, temporally evolving content from a single text prompt is challenging, as models often fail to capture fine-grained temporal cues embedded in language.
As a result, T2V systems tend to focus on the main clause while overlooking temporal subclauses.
For instance, given the prompt \textit{the dog catches the ball and returns it to the man}, the model may generate only \textit{the dog catches the ball}.
This behavior is partly attributable to training objectives that favor short, temporally consistent clips.
To address this limitation, initiated by LVD~\cite{lian2023llm} and other early models like FlowZero~\cite{lu2023flowzero}, FreeBloom~\cite{huang2024free}, and VideoDrafter~\cite{long2024videodrafter}, T2Vs leverage LLMs to decompose prompts into structured scene plans.
These plans guide story construction by drafting keyframes or event sequences.
For instance, EIDT-V~\cite{jagpal2025eidt} utilizes LLMs to generate frame-wise prompts and to identify the scene switch.
Moreover, VideoStudio~\cite{long2024videostudio} and DirectT2V~\cite{hong2024large} ask LLM to convert prompts into multi-scene scripts.
Subsequent event-wise segments can be merged into a coherent video using techniques such as,  noise blending~\cite{henschel2025streamingt2v} or frame interpolation~\cite{liupersonalised, arkhipkin2024improveyourvideos}.

\subsection{Interactivity: Prompt-Generation Alignment}
In the perception–action systems, interactivity characterizes how tightly a world model is coupled with the real world.
In T2V systems, text serves as the primary interface enabling this interactivity.
However, textual expressiveness and visual richness arise from different modalities, meaning that improvements in text modeling do not necessarily translate into better visual generation.
As a result, specialized strategies are required to enhance interactivity in T2V models.

\pb{Prompt Optimization.}
Detailed descriptions, rather than merely expressive prompts, are more effective at aligning generation with visual richness.
However, crafting such detailed prompts is often tedious.
Leveraging their strong language understanding, LLMs provide an efficient solution through prompt rewriting.
For example, POS~\cite{ma2023pos} proposes a semantic-preserving rewriter to prevent semantic drift from na\"ive LLM rewriting.
Runway employs coarse-to-fine progressive refinement instead of passing a long and detailed prompt at once~\cite{saichandran2025progressive}.
VideoTetris~\cite{tian2024videotetris} performs spatiotemporal hierarchical refinement.
MOVAI~\cite{patel2025ai} introduces a compositional scene parser to generate scene graphs with temporal annotations.
InstanceCap~\cite{fan2025instancecap} performs temporal- and attribute-level decomposition with improved chain-of-thought prompting~\cite{wei2022chain}.
MOVi~\cite{rahman2025movi} disentangles the prompt based on the subject.
The rewritten results can further be translated into auxiliary conditioning signals that improve visual control, such as trajectories or layouts~\cite{lu2023flowzero}, reference images or videos~\cite{liupersonalised}, sketches~\cite{liu2025sketchvideo}, depth maps~\cite{xing2024make}, keypoint poses~\cite{bui2025persynth, zhao2025taste, zhang2024fleximo}, and 3D skeletons~\cite{kim2024human}.
Given that grounding is a critical component of vision–text alignment, auxiliary non-textual conditions enhance text–vision grounding by refining cross-modal understanding~\cite{huang2025fine}.
Further, VLMs are commonly employed to reason over the prompt and select appropriate conditioning signals~\cite{li2025bindweave, kim2025free2guide}.

\pb{Visuo-Textual Cross-Attention.}
To integrate text condition to DiT, a common approach is performed by replacing class condition embedding with language model embedding~\cite{chen2024gentron}, and incorporating a visual-text cross-attention layer~\cite{chen2023pixart}.
CogVideoX improves this technique by integrating modality experts that modulate parameters in the adaptive layer norm based on the modality being passed~\cite{yang2024cogvideox}, allowing modalities to be optimized individually yet learned in a union.
Meanwhile, HunyuanVideo proposes a two-stage alignment~\cite{kong2024hunyuanvideo} where each modality is processed individually before being processed as one sequence.
Similarly, two separate weights can be assigned for the two modalities~\cite{esser2024scaling}.

\subsection{Temporal Consistency: Entity}
Entity consistency describes the ability of a model to maintain a stable identity for subjects across frames.
Entity consistency is substantial for preventing visual drift and identity fragmentation over time.


\pb{Reference Representation.}
Identity preservation in T2V refers to maintaining consistent visual characteristics of subjects (e.g., facial features, clothing, and expressions) across frames and viewpoints.
Intuitively, these attributes should remain as close as possible to the initial appearance.
As a result, many methods incorporate reference features during training to anchor subject identity~\cite{tian2024videotetris, liu2025sketchvideo, hong2024large, li2025bindweave}.
ConsistI2V~\cite{ren2024consisti2v} and other models~\cite{xia2024unictrl, khachatryan2023text2video} commonly derive these references from the first frame.
Meanwhile, MEVG~\cite{ni2024ti2v} and other models~\cite{ni2024ti2v, deng2019irc, he2024storytelling} adopt inversion, where the noise initialization is taken from previously generated frames rather than sampled from a Gaussian distribution, thereby reinforcing subject continuity.
Alternatively, some approaches jointly condition the generation process on both the first frame and the last frame to further enhance identity consistency~\cite{yin2025best}.

\pb{Single-Subject Learning.}
Generating a single entity that engages in diverse activities or appears across varying backgrounds remains a significant challenge.
Single-stream generation, which produces all scenes within a single pass, often lacks visual diversity.
In contrast, multi-stream generation synthesizes scenes independently but typically struggles to preserve entity consistency.
To strike a balance between diversity and consistency, diffusion-based attention mechanisms have introduced several strategies.
ShotAdapter~\cite{kara2025shotadapter} and EIDT-V perform transition-aware attention masking that regulates shot-specific prompting.
MoCA~\cite{xie2025moca} leverages cross-attention on mixtures of hierarchical temporal representation~\cite{gal2024breathing}.
ConsisID~\cite{yuan2025identity} decomposes subject representation for high and low-frequency learning. 

\pb{Multi-Subject Learning.}
To ensure consistent generation of multiple subjects within a single video, subject-specific representations must be disentangled to avoid interference with other subjects or the background.
Current T2Vs achieve this form of \say{isolation} through mechanisms like subject-wise attention masking~\cite{tian2024videotetris, wang2024customvideo}.
Additionally, DisenStudio~\cite{chen2024disenstudio} decomposes text features at the subject level, and TIV-Diffusion~\cite{wang2025tiv} integrates subject-specific slot attention modules.

\subsection{Structural Stability: Composition}
Composition encompasses all visual elements in a scene, including foreground subjects and the background.
With this perspective, composition consistency is closely related to view consistency.
The generated world is expected to preserve coherent and stable visual structures across different viewpoints and directions.
This is analogous to a 360$^\circ$ panoramic video, where traversing the scene ultimately leads back to a viewpoint that seamlessly aligns with the original position.

\pb{3D Geometric Prior.}
The 3D geometric representations, such as point clouds, encode rich structural knowledge about the physical world.
From 3D geometry, one can infer static structure, motion dynamics~\cite{wang2025precise}, and lighting directions~\cite{zheng2025vidcraft3}.
Recent studies in T2V have developed several interesting techniques in leveraging 3D prior for structural stability.
GenDef~\cite{wang2023gendef} warps a static reference image in a generative deformation field.
CineMaster~\cite{wang2025cinemaster} facilitates users in providing bounding boxes for scene assembly, while GEN3C~\cite{ren2025gen3c} builds a 3D cache to memorize the structure of previously generated frames.
In addition, Geometry Forcing~\cite{wu2025geometry} enforces angular and scale alignment as auxiliary constraints to align the generated 2D projections with the underlying 3D structure.

\pb{2D Structural Layout.}
Despite its expressiveness, acquiring 3D geometric structure is costly, whether through manual annotation or learned reconstruction models.
To enforce spatial consistency with a more lightweight alternative, recent approaches introduce dynamic 2D layouts across the generated frame sequence.
Common approaches includes incorporating motion trajectories~\cite{liu2024intragen} and spatial displacements~\cite{yatim2024space}.
BlobGEN~\cite{feng2025blobgen} utilizes visual primitive blobs to ground composition.
Dysen-VDM~\cite{fei2024dysen} relies on LLM to generate a scene graph.
MotionBooth introduces a latent shift module to control camera movement. 
Beyond manual specification, models like CAMEO~\cite{nam2025generating} and Direct-A-Video~\cite{yang2024direct} employ a learnable camera module to learn structural cues.
TrajVLM-Gen~\cite{yang2025seeing} employs VLM to predict the trajectory of the next frame.
The frequency domain also offers valuable signals for structural consistency. 
In particular, low-frequency components preserve semantic content and spatiotemporal correlations, thereby maintaining global consistency, while high-frequency components capture motion dynamics.
Accordingly, ConsistI2V~\cite{ren2024consisti2v} utilizes low-frequency information from a reference image as layout guidance. 
ScenarioDiff~\cite{zhang2025scenariodiff} utilizes a mixed-frequency ControlNet where high-frequency components are used to initialize noise.
ByTheWay~\cite{bu2025bytheway} proposes Fourier-based motion enhancement that modulates the high-frequency components of the temporal attention map.
Finally, the learning strategy also plays an important role in structural consistency. 
For example, incorporating localized visual–text cross-attention~\cite{zhang2025tgt} or leveraging shape rigidity losses~\cite{rai2024enhancing} has been shown to further stabilize compositional structure.

\subsection{World Consistency: Physical and Commonsense Conformities}
World consistency is a fundamental requirement for any model that aims to replicate the real world.
It refers to the adherence to the laws of the world, which typically encompass both commonsense constraints and physical principles.

\pb{Learning Physical Law.}
Physical laws are among the most salient indicators of world adherence.
A ball bouncing upon hitting the ground and water flowing faster than milk illustrate fundamental physical regularities.
As a visual generative model, T2V naturally prioritizes visual fidelity and temporal coherence, often neglecting the incorporation of physical knowledge.
As a result, external mechanisms are required to trigger T2V models to respect the physical laws of the real world.
In general, models acquire physical knowledge through three stages: perception of physical properties, simulation of dynamics, and visual rendering~\cite{liu2024physgen}.
Contemporary approaches involve motion cues, such as 3D point trajectories~\cite{chen2025towards} or optical flow~\cite{montanaro2024motioncraft}.
MagicTime~\cite{yuan2025magictime} utilizes time-lapse video for learning metamorphic generation.
VideoREPA~\cite{zhang2025videorepa} distills physics knowledge from pre-trained large video encoders and learning token relation.
NewtonGen~\cite{yuan2025newtongen} relies on physics-informed neural ordinary differential equation to learn Newtonian motions.
Physics learning can also be performed post-training.
For instance, PISA~\cite{li2025pisa} performs fine-tuning by watching stuff drop, and PhysMaster~\cite{ji2025physmaster} leverages reinforcement learning to optimize physics encoder.
Beyond vision-only approaches, VLMs and LLMs can further assist T2Vs in learning physical plausibility.
VLIPP~\cite{yang2025vlipp} leverages physics-aware chain-of-thought~\cite{wei2022chain} reasoning for predicting rough motion.
DiffPhy~\cite{zhang2025think} utilizes LLM to derive physical context from the prompt.
Similarly, WISA~\cite{wang2025wisa} derives the prompt to enrich qualitative physical categories and quantitative physical properties.
GPT4Motion~\cite{lv2023gpt4motion} relies on LLM assistance to generate a script for physical simulators (e.g., Blender).
SGD~\cite{hao2025enhancing} performs counterfactual prompting to suppress physical implausibility.
Moreover, architectural and training strategies, including mixtures of physical experts~\cite{wang2025wisa} and reinforcement learning~\cite{wang2025physcorr}, have been shown to further strengthen the embedding of physical knowledge into T2Vs.

\pb{Commonsense Conformity.}
Beyond physical dynamics, T2V models must also incorporate world commonsense to achieve truly world-like generation.
These include material properties, causal relationships, and factual adherence~\cite{calderaverifyvid}.
Accurate encoding of material properties is essential for maintaining physical plausibility.
For example, a ball landing on a bed should deform and settle rather than bounce as it would on a rigid floor.
Such material-dependent behavior can be modeled by integrating physics-based simulations with material-aware point cloud representations.
PhysCtrl~\cite{wang2025physctrl} proposes a spatiotemporal attention block that emulates particle interactions in point cloud.
PhysChoreo~\cite{zhang2025physchoreo} estimates the static physical properties of all objects in the image through part-aware physical property reconstruction.
In contrast, causal reasoning can be introduced in several ways.
For instance, PhysAnimator~\cite{xie2025physanimator} introduces customizable energy strokes and the incorporation of rigging point support, combined with deformable simulation.
PhyT2V~\cite{xue2025phyt2v} performs chain-of-thought and step-back reasoning to learn causality.

\section{Multimodal Control and Human-In-the-Loop Of T2V}\label{sec:interaction}
\subsection{Leveraging Multimodal Control Through Any-to-Any Generation}
A world model, by definition, must accommodate a wide range of input signals.
In practical scenarios such as gaming, these signals include multi-user controls, such as keyboard commands, haptic inputs, text instructions, and voice signals.
Moreover, the model is expected to support diverse tasks, including video generation, audio synthesis, and haptic feedback.
Such requirements extend the conventional scope of T2V, which is limited to bimodal input and single-task output, toward a more general any-to-any generation paradigm.

\pb{General Framework.}
Exploration of any-to-any generation begins in 2023 with
CoDi~\cite{tang2024any} as one of the pioneering studies.
It emphasizes that the generation capability of the diffusion model is not limited to vision, but to other modalities.
The key enabler is isolated training, where a specific diffusion model is trained for a specific modality.
The input representations from all modalities are projected into the same space before being interpolated for training.
To alleviate the costly all-modal optimization, CoDi utilizes text-any bridging.
After isolated training, CoDi performs joint generation with the assistance of a cross-modal attention layer. 
Following CoDi, there have been numerous developments for diffusion-based any-to-any generation.
For instance, to alleviate the costly training from scratch, modality-specific diffusion models are replaced with a unified modular multimodal diffusion with a joint attention layer~\cite{li2025omniflow}, diffusion latent aligner~\cite{xing2024seeing}, or inter-models inter-layer consistency transfer~\cite{qiu2025multimodal}.

\pb{Multimodal Alignment.}
The alignment complexity in any-to-any generation increases with the number of modalities involved.
Na\"ive multimodal binding often results in poor generation results. 
To tackle this issue, numerous studies have proposed several remedies, such as ensembling modality-specific models and modulating their contribution with learnable routers~\cite{wang2024omnibind}, constructing a multimodal Control-Net to generate compound conditions~\cite{zhang2024c3net}, and binding unified multimodal spaces to an expert single modal space~\cite{wang2024freebind}.
To address modalities with limited data (e.g., haptics), OmniBind leverages distillation from data-rich modality to data-scarce ones, enhancing cross-modal alignment~\cite{lyu2024omnibind}.

\pb{In-Context Understanding From Multimodalities.}
Real-world deployment of a world model demands contextual understanding arising from the integration of multiple modalities. 
For example, text prompts and audio cues in gameplay may carry implicit instructions that require deep contextual comprehension.
Achieving this level of expertise goes beyond simple multimodal alignment, motivating the use of LLMs to enhance reasoning and understanding across modalities.
NExT-GPT~\cite{wu2024next}, CoDi-2~\cite{tang2024codi}, and Vitron~\cite{fei2024vitron} are among the first any-to-any generative models that leverage LLMs.
It projects all modalities into language-like representations and passes them to an LLM that performs semantic understanding and reasoning.
On the output side, multiple strategies exist to enhance the mapping from LLM outputs to generative models.
For instance, UnifiedMLLM~\cite{li2025unifiedmllm} and Spider~\cite{lai2024spider} utilize learnable task tokens and a router, A-Language~\cite{chen2025symbolic} leverages symbolic representation, and ModaVerse~\cite{wang2024modaverse} proposes an I/O alignment strategy where LLM generates meta-responses to instruct the specific generative model.
Finally, a step further from employing LLMs as the alignment engine is to solely utilize autoregressive transformers for multimodal generation~\cite{wang2024worlddreamer, ge2024worldgpt}. 

\subsection{Human In the Generation Loop}
Humans are central to any artificial intelligence system, including T2V. 
In a perception-action system, the world model must align closely with the user’s actual experiences and perceptions. 
This requires human involvement in the generation loop, either by providing feedback on outputs or by interacting with the system during the process.

\pb{Human Feedback.}
Reinforcement learning with human feedback (RLHF) is a \textit{de facto} method for incorporating human feedback into AI models.
Aligning AI generation to human preference not only helps in generating hyper-realistic output but also entails other benefits, such as correcting model reasoning and safeguarding against malicious input triggers.
The customary practice in RLHF is to collect manual human annotation on a small subset of data, which is then used in direct preference optimization (DPO)~\cite{rafailov2023direct} or to train a reward model in proximal policy optimization (PPO)~\cite{schulman2017proximal}.
In T2V, DPO-based RLHF~\cite{xu2024visionreward, zhu2025aligning, liu2025improving} is more commonly employed than PPO-based RLHF~\cite{ahn2024tuning}.
Due to the prohibitive cost of iterative reward optimization, several studies modify DPO through reusing a pre-trained diffusion model as a value estimator~\cite{yang2024using} and recasting the reward optimization as editing task~\cite{yuan2024instructvideo}.
Besides the common DPO and PPO, T2Vs also utilize iterative preference optimization (IPO)~\cite{yang2025ipo, wang2024lift, von2024fabric}.
IPO is an extension of DPO, which leverages a critic model trained with manually annotated human feedback to optimize the generative model iteratively.
It is worth noting that fine-tuning on feedback data also empowers the alignment in terms of physics consistency~\cite{qian2025rdpo}.

\pb{Human-AI Co-Creation.}
Co-creation with a generative model entails active user participation throughout the generation process. 
Rather than limiting user input to the initial text prompt or conditions, co-creation frames T2V as an iterative mechanism that continuously relies on user involvement.
Several works that explore this direction are InteractiveVideo~\cite{zhang2024interactivevideo} that alter the predicted noise with new conditions, IGV~\cite{yu2025position} that leverages four modules supporting the world modeling through user interaction (i.e., control, memory, dynamics, and intelligence), and Imagine360~\cite{wen2025see} that incorporates users' speech-based text prompts.

\section{Datasets and Evaluation}\label{sec:datametric}
\subsection{Datasets}
Datasets evolve following the evolution of T2Vs.
For instance, in the early T2V era, simple datasets such as MovingMNIST~\cite{srivastava2015unsupervised} and PororoSV~\cite{li2019storygan} were frequently used for training.
As the T2V evolved from multi-image generation to video generation and to world modeling, the datasets became more complex.

\pb{General T2V Datasets.}
We observe that there are five most commonly utilized datasets for training T2V: WebVid-10M, HD-VILA-100M, SSV2, MSR-VTT, and LAION-5B.
Webvid-10M is the expansion of WebVid-2M~\cite{bain2021frozen}, which consists of 2.5M pairs of short caption-video from Shutterstock.
The open-domain nature of Webvid-10M encourages learning of diverse visual representations.
However, WebVid-10M contains watermarks~\cite{wang2024worlddreamer}, which challenge the generation of videos with high visual fidelity.
Thus, researchers often supplement WebVid-10M with additional high-quality, watermark-free video data~\cite{guo2023i2v}. 
HD-VILA-100M~\cite{xue2022advancing} consists of 100M pairs of high-resolution videos and long text captions with diverse categories.
Such a characteristic makes HD-VILA-100M suitable for training high-performing T2Vs.
SSV2~\cite{goyal2017something} is a collection of 100k human-object interaction short videos accompanied by natural language template captions as class labels.
The SSV2 represents physical properties of objects and the world through action.
Given the scarcity of high-quality video-text pair data, supplementing the training dataset with an image-text pair dataset like LAION-5B has become a common practice.
The LAION-5B~\cite{schuhmann2022laion} dataset contains a massive collection of image-text pairs.
LAION-5B is the extension of LAION-400M~\cite{schuhmann2021laion}, which curates the data with common crawl and is accompanied by CLIP~\cite{radford2021learning} filtering.
About 50\% of LAION-5B data pairs have English captions.
While containing only image-text pairs, training T2V on LAION provides large-scale semantic grounding and compositional priors of visual representations.

\pb{Complementary World Modeling Datasets.}
General T2V models often face challenges when applied to native world-modeling tasks, such as navigation, autonomous driving, robotics, and gaming. 
A key limitation is the scarcity of relevant samples in standard training datasets. 
To overcome this, researchers typically augment training with domain-specific data.
Several notable representations of these complementary datasets are Room-2-Room, Sekai, nuScene, OpenDV-2K, Ego-Exo4D, OXE, DROID, Atari Grand Challenge, and Platformers.
Room-2-Room~\cite{anderson2018vision} is a visual-language navigation dataset consisting of 90 indoor scenes with 21k navigation instructions.
Given that scenes are derived from panoramic RGB-D data, 
Room-2-Room is capable of assisting the world model in learning structural consistency.
Sekai~\cite{li2025sekai} is an outdoor navigation (e.g., walking and drone-based) dataset with a total of 5k hours of videos with the corresponding text descriptions.
In the autonomous driving domain, nuScene and OpenDV-2K present proper learning sources for scene diversification.
nuScene~\cite{caesar2020nuscenes} comprises of 1k scenes, each with a 360$^\circ$ field-of-view.
The dataset also includes a comprehensive set of sensors, including cameras, radars, and LiDAR, making it a powerful learning resource for world models in ego vehicles.
Following nuScene, OpenDV-2K~\cite{yang2024generalized} proposes an even larger dataset with text description and a more diverse road conditions and driving scenarios.
Meanwhile, in the robotics domain, datasets that cover holistic modalities can be represented by Ego-Exo4D, DROID, and OXE.
Ego-Exo4D~\cite{grauman2024ego} is an ego-exocentric dataset of human activities, totaling 1k hour-long videos.
This multimodal dataset not only includes videos and language descriptions, but also audio, eye gaze, 3D point clouds, pose, and IMU, making it a perfect complement for training a world model.
Meanwhile, DROID~\cite{khazatsky2024droid} and OXE~\cite{o2024open} offer robot activities image-based datasets with 76k and 1M episodes, respectively.
With their broad coverage of tasks, these datasets offer a valuable training ground for world models in robotics.
Moreover, large-scale datasets also exist in the game domain, which is represented by Atari Grand Challenge, Platformers, Yan, and OGameData. 
Atari Grand Challenge~\cite{kurin2017atari} consists of human interaction on Atari 2600 games, consisting of 2k episodes totaling 45 hours of playtime.
Other representative datasets are Platformers~\cite{bruce2024genie} with 6.8M gameplay videos curated from various Internet games, Yan~\cite{ye2025yan} with 400M high-resolution videos and actions from 3D games, and OGameData~\cite{che2024gamegen} with 1M text-video pairs of various games.
Owing to their scale, these gameplay datasets encompass diverse scenarios and player expertise, providing an effective training ground for developing strategic intelligence. 
Lastly, physical simulation datasets~\cite{jiang2025phystwin} can be incorporated to teach the model about objects’ physical properties.

\begin{table}[!htb]
    \centering
    \caption{T2V benchmark datasets for various evaluation components.}
    \label{tab:benchmark}
\begin{tabular}{|c|c|}
\hline
\textbf{Dimension} & \textbf{Benchmark datasets} \\ \hline
General & FETV~\cite{liu2023fetv} || EvalCrafter~\cite{liu2024evalcrafter} || VBench~\cite{huang2024vbench} || HVEval~\cite{wu2025hveval} || T2VEval~\cite{qi2025t2veval}\\ \hline
Compositional & GenAI-Bench~\cite{lin2024evaluating} || T2V-CompBench~\cite{sun2025t2v}\\ \hline
Temporal & TC-Bench~\cite{feng2024tc} || T2VBench~\cite{ji2024t2vbench} || ChronoMagic-bench~\cite{yuan2024chronomagic} || VMBench~\cite{ling2025vmbench} || VideoVerse~\cite{wang2025videoverse} \\ \hline
Safety & T2VSafetyBench~\cite{miao2024t2vsafetybench} || VidProM~\cite{wang2024vidprom} || ViBe~\cite{rawte2025vibe} \\ \hline
World model & WorldSimBench~\cite{qin2024worldsimbench} || PhyGenBench~\cite{meng2024towards} || PAI-Bench~\cite{zhou2025pai} || Physics-IQ~\cite{motamed2025generative} || Morpheus~\cite{zhang2025morpheus}\\ \hline
Reasoning & Gen-ViRe~\cite{liu2025can} || MMGR~\cite{cai2025mmgr} || V-ReasonBench~\cite{luo2025v} || || TiViBench~\cite{chen2025tivibench} || VBench-2.0~\cite{zheng2025vbench}\\
\hline
Others & VC4VG-Bench~\cite{du2025vc4vg} || Q-Eval-100K~\cite{zhang2025q} || StoryEval~\cite{wang2025your}\\ \hline
\end{tabular}
\end{table}

\subsection{Benchmark}
Like the datasets, benchmark datasets have also evolved in tandem with advances in T2V models.
Early benchmark datasets focus on task-specific evaluation, such as human appearance~\cite{yu2023celebv} and simple motion~\cite{hu2023benchmark}.
Holistic benchmark datasets began emerging in 2024. 
They are typically organized into seven assessment categories: general quality, composition, temporal coherence, safety \& hallucination, world modeling, reasoning, and others (Table~\ref{tab:benchmark}). 
General quality remains the most commonly emphasized evaluation criterion.
Benchmarks on compositional quality 
evaluate the relationships of constituents.
Temporal quality benchmarks 
examine the dynamic quality and consistency.
World model benchmark examines how T2Vs conforms to the world rules, including physics.
Meanwhile, reasoning benchmark measure T2V's ability to perform visual reasoning.
Finally, other benchmarks exist for text alignment (VC4VG-Bench~\cite{du2025vc4vg}), human preference (Q-Eval-100K~\cite{zhang2025q}), and story composition (StoryEval~\cite{wang2025your}).

\begin{table}[!htb]
    \small
    \centering
    \caption{Top-performing T2Vs on several benchmark dimensions. \textcolor{blue}{\textit{Blue}} models are open-sourced.}
    \label{tab:score}
\begin{tabular}{|c|m{50pt}|c|c|c|}
\hline
\textbf{Dimension} & \textbf{Benchmark} & \textbf{Component} & \textbf{Top-3 Performer (hi $\xrightarrow{}$ low)} & \textbf{Performance} ($\mu/\sigma$)\\ \hline
\multirow{4}*{General} && Dynamic smoothness & Sora, Runway Gen-3, Dreamina 1.0 & 73.67/10.58\\
& T2VEval~\cite{qi2025t2veval} & Text-video alignment & Sora, Runway Gen-3, Dreamina 1.0 & 72.03/10.28\\
& (Aug 25) & Overall impression & Sora, Runway Gen-3, Dreamina 1.0 & 69.13/11.43\\
&& Commonsense conformity & Sora, Runway Gen-3, Kling 1.0 & 65.63/11.78\\
\hline
\multirow{7}*{Composition} && Interaction & PixVerse-V3, Kling-1.0, Dreamina 1.2 & 74.20/6.40\\
&& Static attribute & PixVerse-V3, Kling-1.0, Dreamina 1.2 & 69.68/0.66\\
& T2V-Comp & Action & PixVerse-V3, Runway Gen-3, \textcolor{blue}{CogVideoX} & 68.11/13.52\\
& Bench~\cite{sun2025t2v} & Spatial & PixVerse-V3, Dreamina 1.2, Kling 1.0 & 58.14/1.22\\
& (Jan 25) & Numeracy & PixVerse-V3, Kling-1.0, Dreamina 1.2 & 49.53/7.87\\
&& Motion & PixVerse-V3, Runway Gen-3, \textcolor{blue}{CogVideoX} & 27.60/0.85\\
&& Dynamic attribute & Runway Gen-3, PixVerse-V3, \textcolor{blue}{Mochi} & 5.19/1.95\\
\hline
\multirow{8}*{Temporal} && Camera control & Veo 3, Minimax Hailuo, Sora 2 & 74.33/1.70\\
&& 2D layout & Minimax Hailuo, Veo 3, Sora 2 & 67.00/0.82\\
& Video & Natural constraint & Veo 3, Sora 2, Minimax Hailuo & 61.67/5.56\\
& Verse~\cite{wang2025videoverse} & Commonsense & Sora 2, Veo 3, Minimax Hailuo & 57.33/4.50\\
& (Oct 25) & 3D depth & Minimax Hailuo, Veo 3, Sora 2 & 54.00/0.00\\
&& Interaction & Sora 2, Veo 3, Minimax Hailuo & 44.00/5.36\\
&& Mechanics & Sora 2, Veo 3, Wan 2.2 & 38.00/4.32\\
&& Material properties & Minimax Hailuo, Veo 3, Sora 2 & 21.33/0.47\\
\hline
\multirow{5}*{World model} && Magnetism & Runway Gen-3, Lumiere, \textcolor{blue}{VideoPoet} & 37.67/6.13\\
& Physics- & Thermodynamics & Lumiere, \textcolor{blue}{VideoPoet}, Runway Gen-3 & 30.00/2.83\\
& IQ~\cite{motamed2025generative} & Solid Mechanics & Lumiere, \textcolor{blue}{VideoPoet}, Runway Gen-3 & 17.67/2.06\\
& (Feb 25) & Optics & \textcolor{blue}{VideoPoet}, Lumiere, \textcolor{blue}{Stable Video Diffusion} & 9.00/2.95\\
&& Fluid dynamics & Lumiere, \textcolor{blue}{VideoPoet}, \textcolor{blue}{Stable Video Diffusion} & 7.50/1.78\\
\hline
\multirow{6}*{Reasoning} && Planning & Hailuo 2.3, Sora 2, Veo 3.1 & 75.60/2.44\\
&& Spatiotemporal & Seedance 1.0, Veo 3.1, Sora 2 & 56.90/3.69\\
& Gen-ViRe~\cite{liu2025can} & Abstract & Sora 2, Hailuo 2.3, Veo 3.1 & 51.27/6.82\\
& (Dec 25) & Analogy & \textcolor{blue}{Wan 2.5}, Sora 2, Hailuo 2.3 & 45.53/5.16\\
&& Logical & Sora 2, Veo 3.1, \textcolor{blue}{Wan 2.5} & 44.47/2.53\\
&& Perceptual & Sora 2, Hailuo 2.3, Veo 3.1 & 43.57/4.58\\
\hline
\multirow{7}*{Storyline} && Easy plotting & Kling-1.0, Hailuo, \textcolor{blue}{Vchitect 2.0} & 54.17/8.08\\
&& Commonsense adherence & Hailuo, Kling-1.0, Pika 1.5 & 36.30/6.78\\
&& Animal story & Kling-1.0, Hailuo, \textcolor{blue}{Vchitect 2.0} & 34.37/10.58\\
& StoryEval~\cite{wang2025your} & Human story & Hailuo, Kling 1.5, \textcolor{blue}{Vchitect 2.0} & 32.30/7.65\\
& (Jul 25) & Object story & Kling-1.0, Hailuo, Pika 1.5 & 28.73/5.99\\
&& Creativeness & Kling-1.0, Pika 1.5, Hailuo & 24.87/7.94\\
&& Hard plotting & Kling-1.0, Hailuo, \textcolor{blue}{CogVideoX} & 10.37/4.68\\
\hline
\end{tabular}
\end{table}


\subsection{Metrics}
Besides the benchmark dataset, evaluation on T2V also incorporates numerical metrics.
There are three general categories of metrics: visual quality, text-vision faithfulness, and temporal dynamics.
The visual quality metric measures the visual fidelity of the generation.
Several widely used visual metrics are IS, FID, and FVD.
Inception Score (IS)~\cite{salimans2016improved} measures the average per-frame KL distance between ground truth label distribution and the generated label distribution from Inception~\cite{szegedy2015going} network.
Fr\'echet Inception Distance (FID)~\cite{heusel2017gans} calculates the average per-frame Fr\'echet distance between the ground truth samples distribution and the generated samples distributions from the Inception-v3~\cite{szegedy2016rethinking} network.
Fr\'echet Video Distance (FVD)~\cite{unterthiner2018towards} measures the maximum mean discrepancy (MMD)~\cite{guo2022attention} between ground truth distribution and generated distribution after being passed to an inflated 3D Inception-v1 network~\cite{carreira2017quo}.
For text-vision faithfulness, some common metrics are CLIP R-Precision and CLIP Score, and some newly proposed scores like VQA Score and TLT Score. 
CLIP R-Precision~\cite{park2021benchmark} is obtained by querying the CLIP model with the generated image and automatically checking how much the retrieved caption matches the ground truth caption.
CLIP Score~\cite{hessel2021clipscore} is calculated by taking the average (CLIP SIM)~\cite{wu2021godiva} or maximum~\cite{han2022show} of cosine similarity between the text and vision embeddings.
VQA Score~\cite{li2024evaluating} measures the probability of a \textit{yes} answer to a confirmation question like \textit{Does this figure show *text*?} from VLM.
TLT Score~\cite{ji2024tltscore} measures the embedding distance between evaluation embeddings with the curated prompt embeddings of the VLM.
Additionally, there are several newly proposed metrics to measure temporal dynamics, such as Dynamic Score~\cite{liao2024evaluation}, MTScore~\cite{yuan2024chronomagic}, CHScore~\cite{yuan2024chronomagic}, and NeuS-V~\cite{sharan2025neuro}.

\section{Applications}\label{sec:application}



\subsection{Metaverse}
With the capability to understand and simulate physical worlds, it is reasonable to assume that T2Vs could facilitate the development of the metaverse, offering a more dynamic, personalized, immersive user experience.
The metaverse, conceived as a collective virtual shared space, merges multiple aspects of digital and augmented realities, including social networks, online gaming, augmented reality (AR), and virtual reality (VR)~\cite{Lee2021AllON}. 
The metaverse thrives on the continuous creation and expansion of its virtual environments and experiences. 
T2Vs can potentially contribute to this aspect by enabling the rapid generation of 3D content to populate virtual worlds, since manually constructing 3D objects is tedious and resource-intensive.
This could include everything from environmental backgrounds and animated textures~\cite{guo2022generating, azadi2023make, xie2024towards} to complex narrative sequences, thereby enriching the diversity and dynamism of the metaverse's content landscape.
Additionally, the potential to rapidly construct virtual 3D objects could open up new possibilities, making feasible what was once thought impossible in the future. 
The T2V advancements suggest the potential for creating digital twins modeled after physical items. 
Other attributes of these items, like sound and tactile feedback (haptics), may be enhanced in addition to a series of images, for the sake of realistic copies of the world model.
T2Vs can also unleash the potential of the metaverse, through studying the interaction between virtual entities and human users. 
This includes evaluating activities that are difficult to carry out in the actual world, such as user studies that may raise ethical issues (e.g., racism or dark patterns~\cite{Wang2023TheDS}) and technological constraints (e.g., deploying a movable 100-meter building~\cite{Delmerico2022SpatialCA}). 
Using the future generation of advanced T2Vs, researchers may simulate augmented reality in virtual environments (i.e., virtual reality) to analyze user behaviors and their response to 3D user interfaces, with the premise that modern virtual reality headsets can provide high-quality video and seamless experiences.



\subsection{Robotics}
In robotic applications, T2Vs can be an affordable platform for robots to learn manipulation actions~\cite{yang2015robot}.
This facilitates open-world learning by lowering the cost and effort for data collection on human demonstration, which was initially performed by video recording of directed choreography with 3D motion capture~\cite{gui2018teaching}.
Early attempts to leverage robot learning via video are made by supplying trajectory signals (e.g., flow vectors) to an action image~\cite{hao2018controllable}.
From this, trajectory learning is improved by incorporating detailed text description~\cite{xu2023controllable} and robot state~\cite{wu2023unleashing}. 
Such a development encourages the works in robot learning to assimilate recent generative AI models such as T2I~\cite{kapelyukh2023dall} and T2V~\cite{du2023video}, which inherit the power of LLM to act as a policy generator or reward for reinforcement learning~\cite{escontrela2024video, sontakke2024roboclip}.
This advancement enables the realization of scalable, unsupervised, and generalizable robot learning.

\subsection{Autonomous Driving}
Similar to the robotic application, T2Vs for autonomous driving are also mainly utilized as data generators for uncommon road scenes, such as traffic accidents~\cite{fang2022cognitive}, which are enormously costly to recreate.
The availability of such data is highly beneficial for designing autonomous vehicles' safety features~\cite{dao2023mm} as it enables learning the trajectory before and after the accident.
Further, video generation for this field can also provide panoramic driving scene data in various road conditions that were originally limited in supply~\cite{wen2023panacea, li2023drivingdiffusion}. 
Output generated by T2Vs enables autonomous vehicle intelligence systems to learn directly from real-world environments instead of being confined in the game environment, which was commonly used to simulate the driving scene~\cite{wang2023drivedreamer, li2024seamless}.
\subsection{Healthcare}
Despite lingering doubts about its trustworthiness, T2V holds the potential to improve the healthcare industry.
Current studies primarily implement these models in healthcare education and medical imaging.
In medical education, T2Vs can help to create training videos for healthcare practitioners whose number is considerably limited since the collection process involves real medical cases~\cite{sallam2024envisioning}.
This is particularly useful in generating educational videos that involve rare cases.
In its function to enhance medical imaging, T2Vs have been explored in various examinations such as radiology~\cite{benaich2023current}, CT scan~\cite{hamamci2023generatect}, and endoscopy~\cite{li2024endora}.
T2V also empowers any-to-any generation in the medical domain~\cite{zhan2024medm2g, molino2025any}.
Besides, T2V may promote the equal distribution of health services between highly developed areas and rural areas.
One interesting study by Loh and Then~\cite{loh2013cardiac} envisions the data conversion of echocardiogram video into text to facilitate a more economical and faster information transmission between urban and rural regions.

\section{Discussion and Future Direction}\label{sec:discussion}

\subsection{Homework on Technical Improvement}
Despite the emergence of high-performing commercial T2Vs and the rapid exploration of many core world model requirements (e.g., consistency and physics), several aspects remain open questions.
First, \textit{diversity and consistency remain two distinct dimensions}.
Generating the whole video at once protects consistency but sacrifices diversity~\cite{zhang2023i2vgen}; on the other hand, multi-step generation tends to prioritize diversity and struggle to maintain consistency~\cite{jagpal2025eidt}.
Second, \textit{consistency is still costly to afford}.
As 3D data, video can be modeled by 3D attention, either full or pseudo.
3D full attention is computationally expensive, while pseudo-3D attention is unstable on scene with large spatial motion~\cite{yang2024cogvideox}.
Some recent advances in temporal attention or autoregressive-based generation have improved the temporal consistency modeling, but still fail to circumvent temporal flickers~\cite{lu2024freelong}. 
Third, \textit{long video synthesis remains challenging}.
Long video generations still rely on a temporal interpolation mechanism whose quality is capped at the base model capacity~\cite{yang2025longlive}.
On the cascaded generation framework, temporal interpolation struggles to model complicated motions~\cite{hur2025high} and is inferior in multi-scene generation~\cite{zhang2023i2vgen}.
Further, developing a true long-video synthesizer is challenged by the train-test data discrepancy, where the model only sees short clips during training but is forced to generalize to long videos at inference~\cite{yin2023nuwa}.
Moreover, training on long videos is not only costly but also limited by the data availability~\cite{yin2023nuwa}.
Fourth, \textit{limited world understanding}.
As discussed in \cref{sec:feature}, current T2Vs still rely on external supplementary cues to understand the world rule, which exposes the insufficiency of world understanding in T2Vs.
Relying on external supplements, however, binds the generation quality to the accuracy of the reference~\cite{hur2025high}.
Full-supervision on world modeling data may be the closest alternative for T2Vs to statistically embed world understanding in the model~\cite{lin2025exploring}; however, it is capped by the availability of world-modeling-specific data for learning~\cite{shao2024360, wang2023lavie}. 
This exhibits the insufficiency of a proper training sample for world modeling in the current pool of training datasets~\cite{jiang2025phystwin}.
Table~\ref{tab:score} reflects these limitations, that even the top-performing modern T2Vs still struggle in physical simulation.
Fifth, \textit{insufficient exploration for flexible interaction}.
Interaction controls remain largely limited to visual and textual forms.
In contrast, true world modeling requires richer modalities (e.g., LiDAR, haptics, and symbolic primitives). 
Further, research on AI-human interaction, such as human-in-the-loop generation, remains limited. 
This highlights the need for further investigation into human-computer interaction (HCI) within T2V~\cite{morris2025hci}.


\subsection{Ethical and Social Implications}
The significant advancement in T2V also entails numerous drawbacks, including, but not limited to, misuse, privacy and copyright, and fairness~\cite{hagendorff2024mapping}.
Hyper-realistic videos from T2V may be misused for creating counterfactual evidence in criminal cases.
The plaintiff may use the generated content in court trials, exacerbating the pressure on the defendant side~\cite{gregory2023fortify}.
Indeed, this calls for adaptation of detection and preventive measures from the lawmakers.
The web data used to train T2V may contain personal identifications whose owners may not intend to share.
If T2V suffers from a data memorization issue, it could generate someone’s face in the video, leaking that individual’s privacy.~\cite{lyu2023pathway}.
Privacy and copyright dilemma is currently faced by WebVid-10M datasets, in which owners of some Shutterstock videos take down their posts, making the videos gradually disappear and ceasing the usefulness of the dataset.
The fairness issue is another caution on T2V as a world model.
For instance, Western culture bias in the training dataset may be inherited by the model, making it generate Western-like output~\cite{gautam2024melting}.
While it sounds trivial, this can be a substantial issue in critical situations, such as robot-assisted surgery or emergency reaction while driving. 
Thus, it is important to pay more attention to the research on model debiasing.

\section{Conclusion}
The arrival of Sora, which claims as world model, has surfaced the importance of a profound understanding of the relationship between text-to-video and world model.
Specifically, it is essential to match world model requirements and the text-to-video generation model's proposals.
Our survey pinpoints that the relationship between the two models is established through the adherence of elements in spatial, action, and strategic intelligence.
Despite moving closer to being a world model, text-to-video generation models still need to solve several important limitations, both from technical and ethical perspectives.
We hope that future studies arising from inter-disciplinary collaborations can solve diverse challenges in synthesizing video from text, to foster the realization of the world model.






\bibliographystyle{ACM-Reference-Format}
\bibliography{bib_mixed,bib_local,chatgpt,bib_text_to_video}





\end{document}